# The Scaffolding Matters More Than the Interface

## A controlled comparison of MCP and CLI tool use across seven agent scaffoldings, five language models, and one software task

Marc Alier Forment[1], María José Casañ Guerrero[1], Francisco José García-Peñalvo[2], Juanan Pereira[3]

[1] Universitat Politècnica de Catalunya (UPC), Barcelona, Spain [2] Universidad de Salamanca (USAL), Salamanca, Spain [3] Universidad del País Vasco / Euskal Herriko Unibertsitatea (UPV/EHU), Donostia-San Sebastián, Spain

marc.alier@upc.edu

---

## Abstract

How much an AI coding agent costs to run can depend more on the agent scaffolding that drives it than on the interface through which it reaches its tools. We set out to measure the cost of tool use over the Model Context Protocol (MCP) against tool use over an ordinary command-line interface (CLI), a difference on which published estimates disagree by more than an order of magnitude while resting on practitioner reports that cannot be reproduced. We ran one fixed software task — six operations against a private online git repository — across seven agent scaffoldings and five language models, and we verified completion by inspecting the repository state rather than trusting the agent's self-report. The dominant effect was the scaffolding. Two of the seven ship no MCP support at all; they completed every run using only the CLI, which shows that MCP is unnecessary for this class of work, and they were 5.0× to 28× cheaper than the five scaffoldings that do support MCP, comparing CLI runs alone with no MCP server attached anywhere. The effect was largest for a small 27-billion-parameter model running locally, whose cost varied 139× across scaffoldings while it completed the task under all of them. The comparison we set out to make proved unstable: thirteen strictly paired MCP-to-CLI ratios span 0.43× to 29×, with outliers on both sides. The two interfaces separate on the cost of failure, where 12.9 per cent of the money spent on MCP runs bought no completed work against 2.2 per cent on CLI runs, but not on its frequency: failures were equally common in both, in the original runs and in their repetitions alike. Agents frequently ignored the interface they were assigned, so comparisons that do not verify actual behaviour measure an unknown mixture. The harness, the task, the verification and the complete dataset are released as open source.

## 1. Introduction

### 1.1 Two mechanisms for tool access

A large language model produces text. To make it useful for software work, something must let it read files, run commands and call remote services. That something is a software harness that the industry is calling the **agent scaffolding**: it takes the user's request, sends it to the model, receives the model's output, executes whatever actions the model asks for, and feeds the results back so the model can continue. Claude Code, OpenAI Codex and qwen-code are widely used examples.

Agent scaffoldings offer capability to the model in one of two ways.

The first way gives the model a structured list of available operations. Each entry in that list, called a **tool schema**, names an operation, describes in words what it does, and specifies the parameters it takes. The mechanism was introduced at scale by OpenAI's function-calling interface [OpenAI 2023], which let a model emit a structured call against a developer-supplied schema instead of free text, and it is usually driven by a reasoning-and-acting loop in which the model alternates between deciding and invoking, as formalised in the ReAct pattern [Yao et al. 2022]. The Model Context Protocol [Anthropic 2024b] standardises how an external service publishes such a list, so that any compatible agent scaffolding can consume it without bespoke integration work. A model that wants to open a pull request selects the `create_pull_request` entry and supplies its arguments. Throughout this paper we call this the **MCP approach**.

The second way, the **CLI approach**, also hands the model a list of tools, but a short one: a shell, a few file operations, and typically one line of prompt noting that programs such as `git` and `gh` are installed. Everything beyond that list rests on what the model already knows about ordinary command-line programs. A model that wants to open a pull request writes `gh pr create --title ... --body ...` and runs it.

The difference between the two approaches is therefore not that one supplies a catalogue of tools and the other supplies none — every scaffolding sends the model some list. The difference lies in where the knowledge sits: the MCP approach carries a description of every operation inside every request, while the CLI approach assumes the operations are already described in the model's weights.

**1.2 Prior estimates of the cost difference**

Every tool schema occupies space in the model's context window, and the MCP catalogue must be sent again with every request, because a language model retains nothing between requests. An agent that takes twenty exchanges to finish a task therefore transmits and pays for that catalogue twenty times. The command-line approach sends one schema no matter how many different programs the model eventually invokes, because the knowledge of what `gh` can do already resides in the model's weights rather than in the request.

This reasoning is sound, and it has produced several published estimates of the penalty. The most widely circulated is a practitioner benchmark reporting that MCP servers consume 35 times more tokens than equivalent command-line tools [MindStudio 2026]; two further practitioner reports put the penalty lower and attribute it to the size of the MCP catalogue and the number of calls made [Reinhard 2026; Vensas 2026].

These figures should be read with three qualifications. None is peer-reviewed; all three originate in the same practitioner community; and each was produced with a different agent scaffolding on a different workload, with the measurement conditions reported too loosely to reproduce. The 35× figure in particular is frequently cited as though it were a stable property of the protocol, when its own authors attribute it to large MCP catalogues and high call counts. Taken together the reported range spans roughly 3× to 35×, which is wide enough that the underlying quantity is evidently sensitive to something the reports do not isolate. Section 4 identifies one candidate.

A second argument concerns not the size of the request but the model's familiarity with what is in it. Steinberger [2026] puts it as an asymmetry in training data: the operation of a Unix shell is represented throughout the pre-training corpus of every current model — manual pages, tutorials, source repositories, decades of questions and answers — whereas the syntax of tool schemas is comparatively recent and enters mainly through post-training. On this account a new command-line tool is "just another Unix command," discovered at need by reading its `--help` output, while MCP requires the model to follow a convention it has seen far less often.

If that asymmetry is real, its effects should be most visible in smaller models, which have less capacity to spare for conventions thinly represented in what they learned. Section 6 examines that case through `qwen3.6-27b`, a dense model of 27 billion parameters — dense meaning every parameter takes part in every token, so the headline figure is also the working size [vLLM 2026]. For scale, the largest model in the same family at the time of writing, `Qwen3.8-Max`, is a sparse mixture-of-experts model of 2.4 trillion parameters [Alibaba Cloud 2026]; it took no part in this study. Between the model we ran on one workstation and its vendor's flagship there is a factor of about ninety in declared size.

**1.3 Study design and contributions**

The disagreement described above has a structural cause: the published estimates were each produced with a different agent scaffolding, on a different task, and none reports the conditions precisely enough for another party to reproduce it. A figure obtained by one team with one scaffolding on one workload has been read as a property of the protocol, which it is not.

Our design responds to that directly. We hold the task fixed — the same six operations, the same repository, the same verification — and vary three things independently: the agent scaffolding, the language model, and the tool interface. Holding the task constant makes the scaffoldings comparable to one another; varying the interface within each scaffolding separates the cost of the interface from the cost of the scaffolding carrying it; and including two scaffoldings that have no MCP support at all provides a reference point that no within-scaffolding comparison can supply.

Two further choices matter. We verify completion by inspecting the repository rather than by accepting the agent's report, because an agent that believes it succeeded and an agent that did succeed produce identical prose. And we record which tools were actually called, rather than which were made available, because Section 7 shows those differ often enough to invalidate the assumption that an agent uses what it is given.

Because the estimates this study responds to are not reproducible, the whole apparatus is released rather than described: the measurement harness, the task definition, the verification code and the complete run-by-run dataset are published at `github.com/Lamb-Project/mcp-vs-cli-bench` and archived at Zenodo as version 1.0.0, DOI 10.5281/zenodo.21851992 [Alier et al. 2026b]. Every figure and table below can be regenerated from that repository, and every claim checked or contradicted with the instruments that produced it.

The study also tests a design hypothesis from our earlier methodology work. *Agents All the Way Down* [Alier et al. 2026a] argues that a custom agent — built for one job, its task boundary known in advance — should shed the general-purpose infrastructure it was prototyped with, and be harvested into a small command-line program. That prescription predicts that carrying general-purpose tooling into a well-specified task has a measurable running cost. This experiment was designed to measure one such cost, the MCP catalogue's. We fixed the task and its verification before running the matrix, and we committed to publishing the results whatever they showed. They did not show what we designed for: the MCP comparison came out inconclusive, and the larger, more stable effect sat with the scaffolding itself.

The paper proceeds as follows. Section 2 describes the method, including how cost is conditioned on completion and what the study itself cost to produce. Section 3 reports what the experiment cost on both of its **arms** — the runs made with MCP attached and the runs made with only the CLI — across the seven scaffoldings. Section 4 examines the cost of the MCP interface within scaffoldings that support both — the paired comparison, the effect of caching, the delivery method, and the cost of failure runs. Section 5 reports cost by model, and Section 6 analyses the performance of open-weight models on local hardware. Section 7 reports how often agents followed instructions and actually used the interface they were assigned. Section 8 sets out the limits of these results, including a replication that measures run-to-run variation, and Section 9 discusses what follows.

## 2. Method

Our goal was to design a task that the tested agent could complete in one single run. The task had to involve several tool calls; it had to be complex enough to be significant for the study; we needed a way to verify whether the agent had actually achieved the desired result; and it had to be achievable for mid-2026 agents. We went through three iterations of the design. The first, which involved four tool calls, proved far too easy. The second proved impossible for most of the agents we tried it on. The final design proved to be the sweet spot.

### 2.1 The task

Every run performed the same six operations against an online private git repository, hosted on github. The task required the model to:

1. Locate a specific open issue.
2. Create a branch.
3. Apply a supplied patch to a file.
4. Commit the change.
5. Open a pull request from that branch.
6. Report how many files a named directory contains.

We chose this task because a capable model can finish it in a single session, it requires both local file work and remote service calls, and every step leaves a durable trace that can be checked afterwards.

### 2.2 Verification of completion

We did not ask the agent whether it had succeeded. After every run we queried the repository through the GitHub API and checked four conditions independently: whether the branch existed, whether the file contained the patched content, whether a pull request had been opened, and whether the reported file count was correct. **Completion** throughout this paper means the percentage of those four conditions satisfied. A run that reports success but leaves no branch behind scores nothing for that condition.

The repository was restored to its starting state before every run, so that work left behind by one run could not be credited to the next.

### 2.3 The seven agent scaffoldings

For our experiment we selected the seven agent scaffoldings below. Four of the five most-starred open-source agents on GitHub are here, alongside two that ship no MCP support at all and serve as controls; Section 8 sets out the selection criterion and what it leaves out.

| Scaffolding | Version | Can use MCP catalogues |
|---|---|---|
| Claude Code | 2.1.220 | yes |
| OpenAI Codex | 0.146.0 | yes |
| qwen-code | 0.21.2 | yes |
| Hermes | 2026.8 | yes |
| opencode | 1.18.15 | yes |
| pi | 0.73.1 | **no** |
| Tau | 0.3.6 | **no** |

The first five can attach an MCP catalogue and were tested both ways. The last two provide no MCP support at all and were tested with a command line only. They are included as a control: they show what the task costs when MCP is not merely unused but architecturally unavailable.

Tau is an independent reimplementation, in Python, of the design philosophy behind pi. Because the two were written separately, by different authors, in different languages, while pursuing the same deliberately minimal design, close agreement between them indicates that a result follows from that design rather than from either scaffolding's particular implementation.

### 2.4 Experimental arms

Each agent scaffolding capable of both was run under two conditions, which we call **arms**, following clinical-trial usage:

- In the **MCP arm**, the official GitHub MCP server was attached. It publishes forty-four tools covering issues, branches, commits and pull requests.
- In the **command-line arm**, no MCP server was attached, and the agent had its shell and the `gh` command-line client.

The arms were isolated so that an agent could not quietly use the other one. In the command-line arm no MCP server was configured at all. In the MCP arm the MCP server held its own credential, while the agent's shell was pointed at an empty configuration directory containing no credentials, so that any attempt to fall back on `gh` failed visibly rather than succeeding unnoticed. Section 7 explains why this isolation proved necessary.

### 2.5 Instrumentation and measures

Every request from every agent scaffolding was routed through a local proxy server, which recorded for each request:

- **Input tokens —** the quantity of text sent to the model. This is the figure MCP inflates, and it is the cost measure used throughout this paper.
- **Cached tokens —** the portion of that input the provider recognised from an earlier request and billed at a reduced rate.

- **Tool calls —** the name of each operation the model invoked.
- **Schemas per request —** how many tool descriptions the agent scaffolding included in the request.

Routing requests through a single proxy matters because some agent scaffoldings delegate work to sub-agents, whose consumption would otherwise go unrecorded and make an expensive scaffolding appear cheap.

**One credential is an exception.** A subscription account authenticates against its own vendor and nowhere else, so the two cells we ran that way bypass the proxy and report their own usage. The scaffolding carrying that credential is not so restricted: Claude Code takes an environment variable naming the endpoint it should call, and our proxy serves an endpoint of the required shape, so every other Claude Code cell is measured the same way as every other scaffolding. The limit belongs to the account, not to the software.

Where both figures exist they agree exactly. Across the twelve proxied Claude Code runs the scaffolding's own totals match the proxy's to the token, in both directions of the comparison and on runs of a quarter of a million tokens. A self-reported number cannot check itself; the exact agreement on the proxied runs is why we trust the two cells that only self-report.

Five language models were used: two hosted by OpenAI, one by Anthropic, and two open-weight models served locally on our own hardware. Not every scaffolding was run against every model; the coverage of each is given in Table 1.

Every dollar figure in this paper is a modelled cost, not a bill. We price each run at the public per-token list rates published by OpenRouter (`openrouter.ai`), captured on 3 August 2026, and we apply one price list uniformly to every cell — including the two locally served models, which are priced counterfactually, as what renting equivalent capacity would cost. Cached input is billed as input, with the cache share reported as its own column, so a reader can apply any provider's discount; Table 4 shows one such sensitivity. Real spending is reported separately, in Section 2.10, and never mixed into these columns.

### 2.6 Two corrections to the measurement

Our verification originally checked five conditions rather than four. The fifth compared each pull request against a particular issue number, but the routine that reset the repository between runs assigned a new issue number every time, so no run could ever satisfy it. All completion figures in this paper are computed over the four conditions that are actually satisfiable.

The second correction was a configuration error of our own. In one scaffolding's configuration file the MCP server's credential was written as a variable name that the scaffolding expands only in a bracketed form, so the server started with the name of the credential in place of its value and was refused by the API on every call it made.

Nothing about the failed run looked wrong from outside. The server was attached, the MCP catalogue was transmitted in full, the agent called MCP tools, and it finished by reporting what it had managed to do — a partial completion, an ordinary-looking cost, no error anywhere in the record. Read from outside it was indistinguishable from a run in which the protocol had not been enough. Only the agent's own closing message, which named the unexpanded variable, said otherwise.

**When a run failed because of an error in our setup or configuration, we fixed the error and repeated the run.** Counting such a run as a failure would have blamed the protocol for our mistake:

a rejected credential measures the harness, not the interface under test. Here we fixed the credential, verified against the API that the server now authenticates, and repeated the affected cell three times. It completes on all three, using MCP tools throughout.

The correction matters because the cell was the most expensive in the MCP arm. Counted as a failure it carried a quarter of all spending on that arm into the wasted-cost share of Section 4.4, which stood above forty per cent before this correction and stands at 12.9 with the cell measured properly. Nothing about the protocol changed in between. An instrument of this kind does not fail by crashing; it fails by returning a number one would have believed.

### 2.7 Construction of the measurement harness

The measurement harness is itself an instance of the method it measures; Section 3.2 uses that fact.

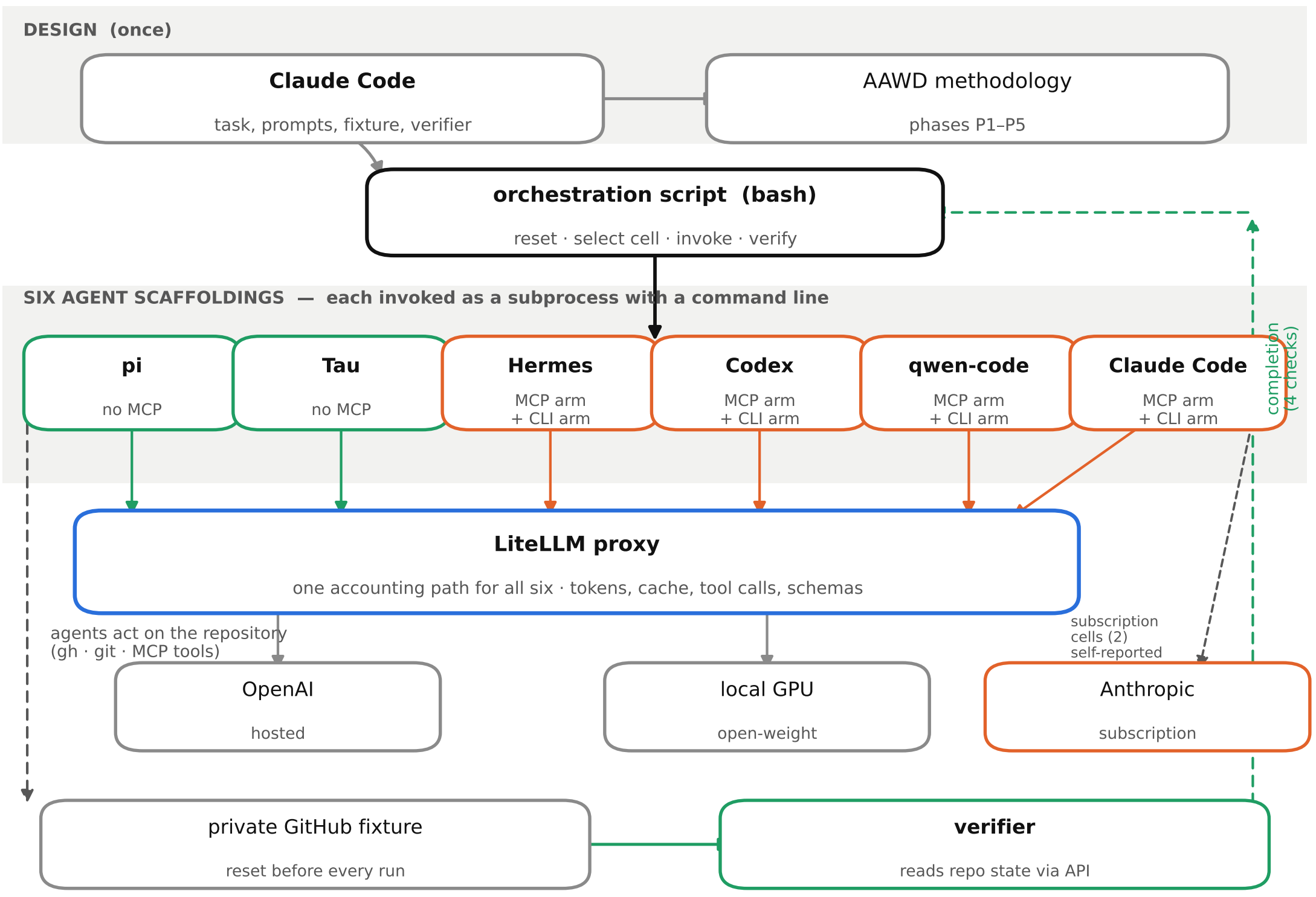

*Claude Code appears twice: it was used in the design and orchestration of the study, and is also one of the six scaffoldings measured by it. All six cross the proxy. Only the two cells run on a subscription account bypass it, because that credential authenticates to Anthropic alone.*

Figure 1: The measurement harness. A single shell script resets the fixture, selects a scaffolding and a model, invokes the agent as a subprocess, and calls the verifier. Every scaffolding is reached identically, through a command line, and every one is measured at the LiteLLM proxy, so none can under-report by delegating work to a sub-agent. The exception is the credential rather than the scaffolding: the two cells run on a subscription account reach the vendor directly and report their own usage. Green marks the two scaffoldings with no MCP client.

We built it following the agent-construction methodology set out in *Agents All the Way Down* [Alier et al. 2026a]. A general-purpose assistant was used for the design work — specifying the task, drafting the prompts, preparing the repository fixture and writing the verification code — and the resulting agents were then run through the seven scaffoldings under test. Orchestration is a shell script: it resets the fixture, selects a scaffolding and a model, invokes the agent as a subprocess, collects its output, and calls the verifier.

That orchestration illustrates the very property this study measures. Every scaffolding tested exposes a command-line entry point, which is what makes driving seven of them from one script possible at all; there is no comparable way to operate seven different agents through their MCP catalogues from a shell loop. The composability that lets an agent be treated as an ordinary program — invoked, piped, scripted, retried — is the same composability that lets an agent use ordinary programs. We depended on the first in order to measure the second.

### 2.8 How cost is reported

Cost and completion answer different questions, and this paper does not mix them.

A run that fails returns no completed work in exchange for the tokens it consumed, so its cost per unit of delivered work is unbounded. Averaging such a run together with a successful one produces a figure that describes neither.

Failure is the expensive outcome. In this study the median failed run consumed **181,806 input tokens against 81,510 for the median completed run** — more than twice as much for nothing. Every failed run cost between 4.7 and 22.9 times a completed run on a scaffolding with no MCP client. An agent that fails does not stop early and cheaply; it works hard, spends more than a successful agent would, and delivers nothing.

Three rules follow, applied throughout.

**Cost figures are computed over runs that completed the task.** The question “what did this job cost” has no answer where the job was not done.

**Completion is reported separately**, as its own quantity, never folded into a cost.

**Where a condition produced no completed run, no cost is reported for it.** The cell is marked undefined rather than imputed, and Section 4.1 shows that such absences are informative in their own right.

Section 4.4 then reports the cost of failure directly, over every run and with nothing excluded.

### 2.9 Repetition

The matrix described above runs each configuration once. To establish how far a single run can be trusted, every configuration using a locally served model was additionally run **three times**, with nothing changed between repetitions. Local inference carries no marginal cost, so the price of this was time rather than money; the hosted configurations were not repeated, for budget reasons.

That produced a second dataset of 90 runs, of which 70 executed and 61 completed the task. It is published alongside the main matrix, and it is used for two purposes: to measure how much an identical configuration varies between runs, which sets the resolution limit of the whole study, and to test whether the principal result reproduces on data gathered after the analysis was fixed. Both are reported in Section 8.1.

Unless stated otherwise, every figure and table in Sections 3 to 7 comes from the main matrix of 54 cells, not from the repetitions.

### 2.10 Disclosure: the cost of producing this study

The harness described in Section 2.7 was built and operated with an agent, so the study has a production cost of its own, and a paper about what agents cost should not be silent about what it cost.

The work ran over eight days in August 2026 across two Claude Code sessions: a six-day session of 2,222 assistant turns that built the harness and ran the matrix, and a second session of 1,388 turns that measured the sixth and seventh scaffoldings properly and revised the analysis around them. Aggregated across both:

| Quantity | Tokens |
|---|---:|
| Input, read from cache | 1,612,245,965 |
| Input, written to cache | 48,680,662 |
| Input, uncached | 6,805 |
| **Total input** | **1,660,933,432** |
| Output | 3,798,219 |

Three features of this profile connect to results reported later.

**The cache carried 97.1 per cent of the input.** An agent session is a conversation that grows monotonically: every turn re-sends the whole history plus whatever the last tool call returned. Without caching the arithmetic is prohibitive; with it, the marginal cost of a turn is roughly the new material alone. This is the same mechanism examined in Section 4.2, observed at a scale three orders of magnitude beyond any individual benchmark run.

**Input exceeded output by 437 to 1.** Agentic work is overwhelmingly a matter of reading — files, command output, prior turns — rather than generating. Cost models built around output length will misprice it badly.

**Uncached input was 6,805 tokens across eight days**, four ten-thousandths of one per cent of the total. Essentially every token these sessions paid for had been seen before in the same conversation.

Three caveats. These figures cover the interactive sessions that designed the harness and analysed the results; they exclude the benchmark runs themselves, which are reported throughout the paper and were measured separately at the proxy. The sessions were billed under a subscription rather than per token, so no dollar figure is given: pricing them at list rates would describe a transaction that did not occur. And a session measured from inside itself can only ever report a snapshot: the second session was still running when these totals were taken.

## 3. Results: cost and reliability across scaffoldings

Figure 2 and Table 1 report what the same task cost under each of the seven agent scaffoldings. Cost is the median input-token consumption of a **completed** run, following the rule that Section 2.8 sets out, and it is aggregated across all five models and both arms — the runs with MCP and the runs with only the CLI. Completion is shown separately, over all runs attempted. We report medians because a small number of runs consumed very large amounts and would otherwise dominate an average.

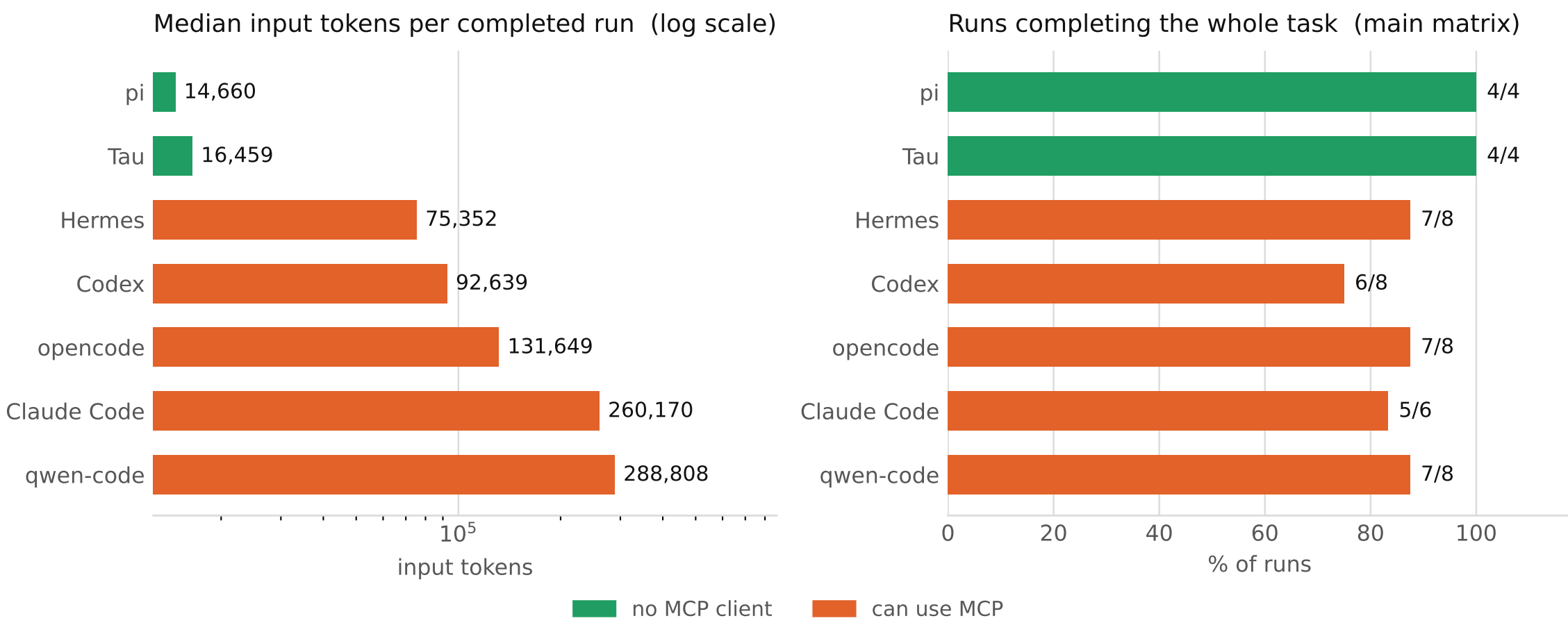

Figure 2: Cost and reliability of the seven agent scaffoldings, over the 54-cell main matrix. The left panel gives median input tokens per completed run on a logarithmic scale; the right panel gives the percentage of runs satisfying all four verification conditions. Green marks the two scaffoldings with no MCP client. The completion panel covers this matrix only: in the separate replication of Section 8.1, Tau failed one of twelve further attempts, so its record across both datasets is nineteen completions in twenty.

**Table 1.** Cost and reliability by agent scaffolding, over runs that completed the task. "MCP" records whether the scaffolding ships a client at all. "Completed" gives completed runs out of runs attempted. "Input" and "Output" are median token counts for a completed run, and "Relative" expresses input as a multiple of the cheapest scaffolding. "Cost" is the median modelled cost of a completed run, priced at each run's own model rate (Section 2.5); the model mix behind each row differs, so compare rows by tokens, and see Table 7 for cost by model. Completion counts are for this matrix; Section 8.1 reports a separate replication in which Tau failed once in twelve further attempts.

| Scaffolding | MCP | Completed | Input | Output | Relative | Cost |
|---|---|---|---|---|---|---|
| pi | no | 4/4 | 14,660 | 824 | ×1.0 | $0.0124 |
| Tau | no | 4/4 | 16,459 | 1,064 | ×1.1 | $0.0149 |
| Hermes | yes | 7/8 | 75,352 | 991 | ×5.1 | $0.0581 |
| Codex | yes | 6/8 | 92,639 | 1,226 | ×6.3 | $0.0838 |
| opencode | yes | 7/8 | 131,649 | 1,186 | ×9.0 | $0.0909 |
| Claude Code | yes | 5/6 | 260,170 | 1,197 | ×17.7 | $0.3347 |
| qwen-code | yes | 7/8 | 288,808 | 1,367 | ×19.7 | $0.1230 |

The task, the models and the verification were identical in every row; only the scaffolding driving the model changed. Between the cheapest and the most expensive lies a factor of **20**. No comparison between tool interfaces reported later in this paper approaches that magnitude.

This is the paper's principal finding, and it rests on the largest sample we have: forty completed runs, of forty-six executed, spread across all seven scaffoldings and all five models. When the running cost of an agent is at issue, the scaffolding chosen to drive the model is a larger factor than the interface through which that model is given tools.

Cost and reliability moved together for the cheapest scaffoldings and diverged for the expensive ones. The two cheapest scaffoldings, pi and Tau, completed every run they attempted in this matrix; the replication of Section 8.1 qualifies that with a single failure by Tau in twelve further attempts, on the smallest model. Among the five that support MCP, failure does not track cost: Codex failed a quarter of its runs while costing a third of what qwen-code did. What holds is the weaker statement — the scaffoldings that never failed are also the cheapest — and not a monotonic relation between price and reliability.

pi and Tau agree closely: 14,660 and 16,459 median tokens, a difference of twelve per cent, both completing every run. Since the two scaffoldings share a design philosophy but no code, this agreement points to the design as the cause rather than to either implementation.

The effect was largest for the smallest model we tested, which runs contrary to the common expectation that weaker models need more elaborate support. A 27-billion-parameter open-weight model completed the task under Tau while consuming 17,416 tokens. The same model, on the same task, under Codex with the MCP server attached, consumed **2,418,828 tokens** across 108 tool calls — 139 times as much — and reached completion only after considerable repetition. The model was capable of the work in both cases. What differed was how much the surrounding scaffolding made it spend to get there.

### 3.1 Why input carries the comparison

Output is reported in Table 1 alongside input. It varies from 824 to 1,367 tokens across the seven scaffoldings — a factor of 1.7 — while input varies by a factor of 20. The task demands very nearly the same amount of generation whichever scaffolding performs it; what differs by an order of magnitude is how much context each carries in order to produce it.

That is why the analysis is expressed in input tokens. Output is recorded for every run and published with the dataset, but it is close to constant and therefore explains almost none of the difference between scaffoldings.

Output nevertheless matters to the bill, because providers charge several times more for it. Its share of the modelled cost of a completed run ranges from **24 per cent for pi and 25 per cent for Tau down to 2.9 per cent for qwen-code and 2.6 per cent for Claude Code**. The pattern is the arithmetic consequence of the first: where input is small, output is a substantial fraction of the total; where input is large, it is negligible. A practitioner economising on output length is addressing a quarter of the cost on a minimal scaffolding and under a fiftieth of it on a verbose one.

### 3.2 Comparison at a fixed interface

Table 1 aggregates both arms, which mixes runs that had an MCP server attached with runs that did not. A sharper comparison uses only the command-line arm, where **no MCP server was attached to any scaffolding**. Every run in Table 2 used ordinary command-line tools; the only difference between the rows is which scaffolding drove the model.

**Table 2.** Command-line arm only. No MCP catalogue was attached in any of these runs. "MCP client" records whether the scaffolding ships one at all. "Runs" counts completed runs, per the rule of Section 2.8. "Cache" is the median share of input tokens billed at the reduced cached rate. "Cost"

is a median over each row's own model mix, per the Table 1 caveat; Claude Code's rests on two completed runs, one of them on the dearest model in the study.

| Scaffolding | MCP client | Runs | Tokens | vs pi | Cache | Cost |
|---|---|---|---|---|---|---|
| pi | **no** | 4 | 14,660 | ×1.0 | 88% | $0.0124 |
| Tau | **no** | 4 | 16,459 | ×1.1 | 77% | $0.0149 |
| Codex | yes | 4 | 82,378 | ×5.6 | 80% | $0.0352 |
| Hermes | yes | 3 | 83,954 | ×5.7 | 93% | $0.0922 |
| opencode | yes | 3 | 137,800 | ×9.4 | 93% | $0.0909 |
| qwen-code | yes | 4 | 297,649 | ×20.3 | 88% | $0.1666 |
| Claude Code | yes | 2 | 410,797 | ×28.0 | 98% | $0.7251 |

Two conclusions follow.

**An MCP catalogue is not necessary for this work.** pi and Tau have no MCP client and cannot acquire one. They completed every run using `git`, `gh` and a shell. Whatever MCP contributes on a task of this kind, it is not access: the same operations were reachable through command-line tools that already existed. This point generalises beyond our task, since mature command-line clients exist for many of the services that also publish MCP servers — `playwright` and the Playwright MCP server expose substantially the same browser operations.

**Scaffoldings without an MCP client were far cheaper than widely used ones carrying the same workload.** Between pi and Codex there is a factor of 5.6, and between pi and Claude Code a factor of 28, with no MCP catalogue attached in any of those runs. The high cache rates across the whole table — between 77 and 98 per cent — show this is not an artefact of one scaffolding caching better than another.

What we cannot do is attribute the difference to MCP support specifically. Our sample contains two scaffoldings without an MCP client and four with one, and they differ in other respects we did not measure. The finding is a comparison between two populations of scaffolding, not an isolation of one feature. Section 4 measures MCP's cost directly, within scaffoldings that support both interfaces, which is the comparison that does isolate it.

## 4. Results: the cost of MCP

### 4.1 Within-scaffolding comparison of the two arms

Figure 3 and Table 3 compare the two arms within each scaffolding that supports both — its runs with MCP attached against its runs with only the CLI — over runs that completed the task. The ratio divides the MCP arm's median by the command-line arm's, so a value above 1 means MCP cost more.

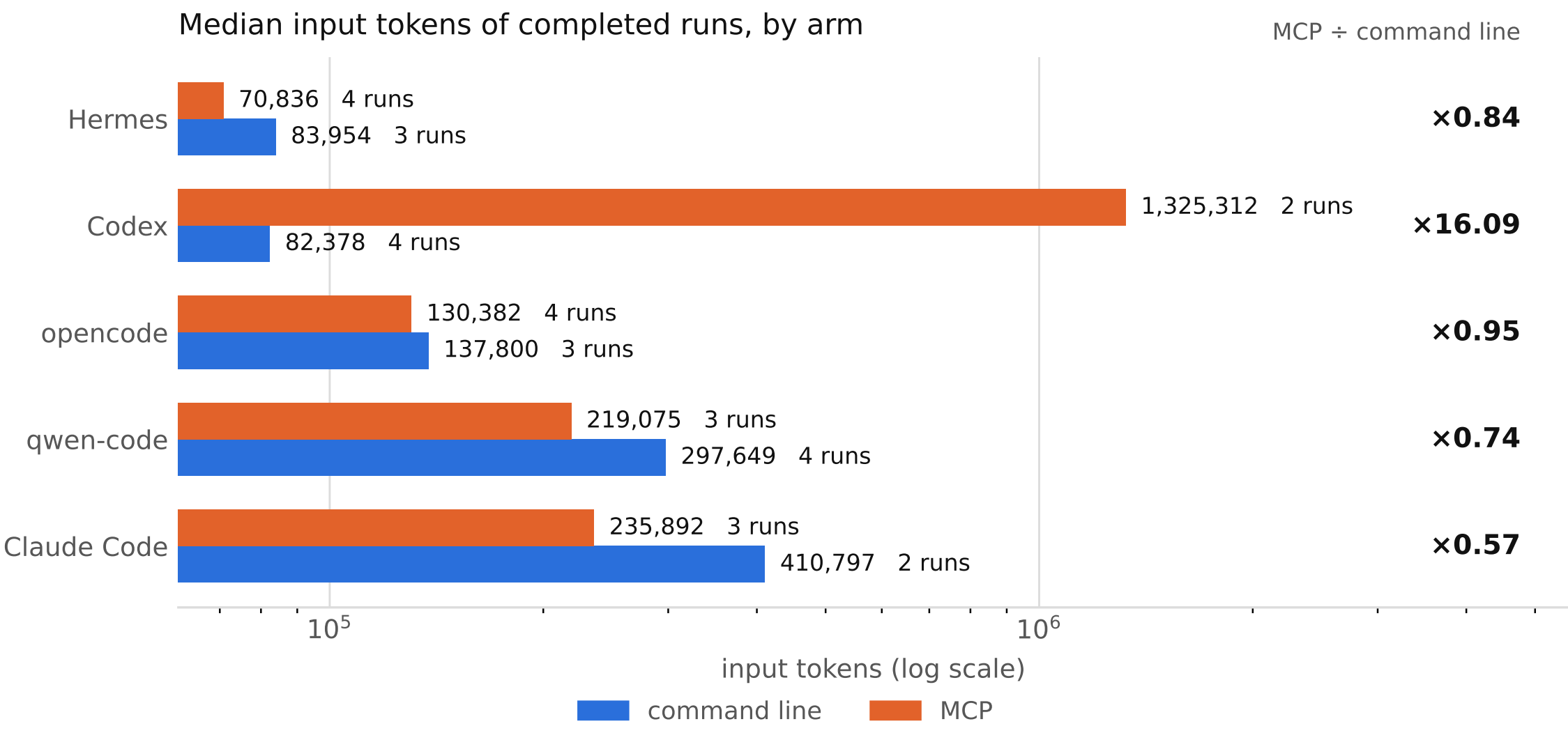

Figure 3: Median input tokens of completed runs in each arm, within each MCP-capable scaffolding. Blue is the command-line arm, orange the MCP arm, on a logarithmic scale; the run counts beside each bar are completed runs in each arm.

| Scaffolding | Command line | MCP | Ratio | Runs (cmd / MCP) |
|---|---|---|---|---|
| Claude Code | 410,797 | 235,892 | ×0.57 | 2 / 3 |
| qwen-code | 297,649 | 219,075 | ×0.74 | 4 / 3 |
| Hermes | 83,954 | 70,836 | ×0.84 | 3 / 4 |
| opencode | 137,800 | 130,382 | ×0.95 | 3 / 4 |
| Codex | 82,378 | 1,325,312 | ×16.09 | 4 / 2 |

**Table 3.** The two arms within each MCP-capable scaffolding, completed runs only. “Runs” gives how many runs in each arm completed. Each arm’s median covers whichever models completed in that arm, so the two sides of a ratio can rest on different model mixes; the thirteen strictly paired ratios in the text are the like-for-like comparison.

The result is not a consistent penalty in either direction. Codex cost 16 times more with MCP attached; Hermes cost 16 per cent less; qwen-code 26 per cent less; Claude Code 43 per cent less; opencode was within five per cent of parity. Restricting to the thirteen cells where a single scaffolding ran a single model in both arms and completed both, the individual ratios span **0.43 to 29.06**, with a median of 0.93.

We therefore report the conventional within-scaffolding comparison as **inconclusive on this evidence**. Thirteen paired observations, spanning a factor of nearly seventy, against the run-to-run variation measured in Section 8.1, cannot resolve an effect of the tool interface in either direction.

That is itself a useful result. The published estimates in Section 1.2 disagree by more than an order of magnitude, and a bimodal distribution of this shape is what would produce exactly that disagreement: each study samples a few points and reports one of them as the value. The quantity is not stable enough to have a single value.

### 4.2 Cost after caching

Providers bill repeated context at a reduced rate, and an MCP catalogue is identical on every request, so it should cache well. Table 4 prices the same completed runs twice: once at list rates, and once applying a ninety per cent discount to cached input, which is close to what the hosted providers in this study charge.

| Arm | Runs | Tokens | Cost, list price | Cost, cached discounted |
|---|---|---|---|---|
| Command line | 16 | 119,916 | $0.0934 | $0.0287 |
| MCP | 16 | 187,954 | $0.1044 | $0.0345 |

**Table 4.** Cost of the two arms under two billing assumptions, completed runs only, across the five MCP-capable scaffoldings. Local models are priced counterfactually throughout, as explained in Section 2.5.

The two arms are close in cost under either assumption: the MCP arm consumed twice the tokens and cost about ten per cent more at list price, or half as much again once cached input is discounted. Neither margin means much, because the completed runs in the two arms are not drawn from the same mix of models, and the per-token prices of those models differ by a factor of twenty. It is a further reason to treat the pooled within-scaffolding comparison as uninformative, and to prefer the paired comparison of Section 4.1.

### 4.3 Catalogue delivery method

Part of the spread in Table 3 has a measurable cause. The proxy recorded how many tool schemas each scaffolding included in each request, and that number separates the one scaffolding whose MCP arm was cheap from those whose MCP arm was expensive.

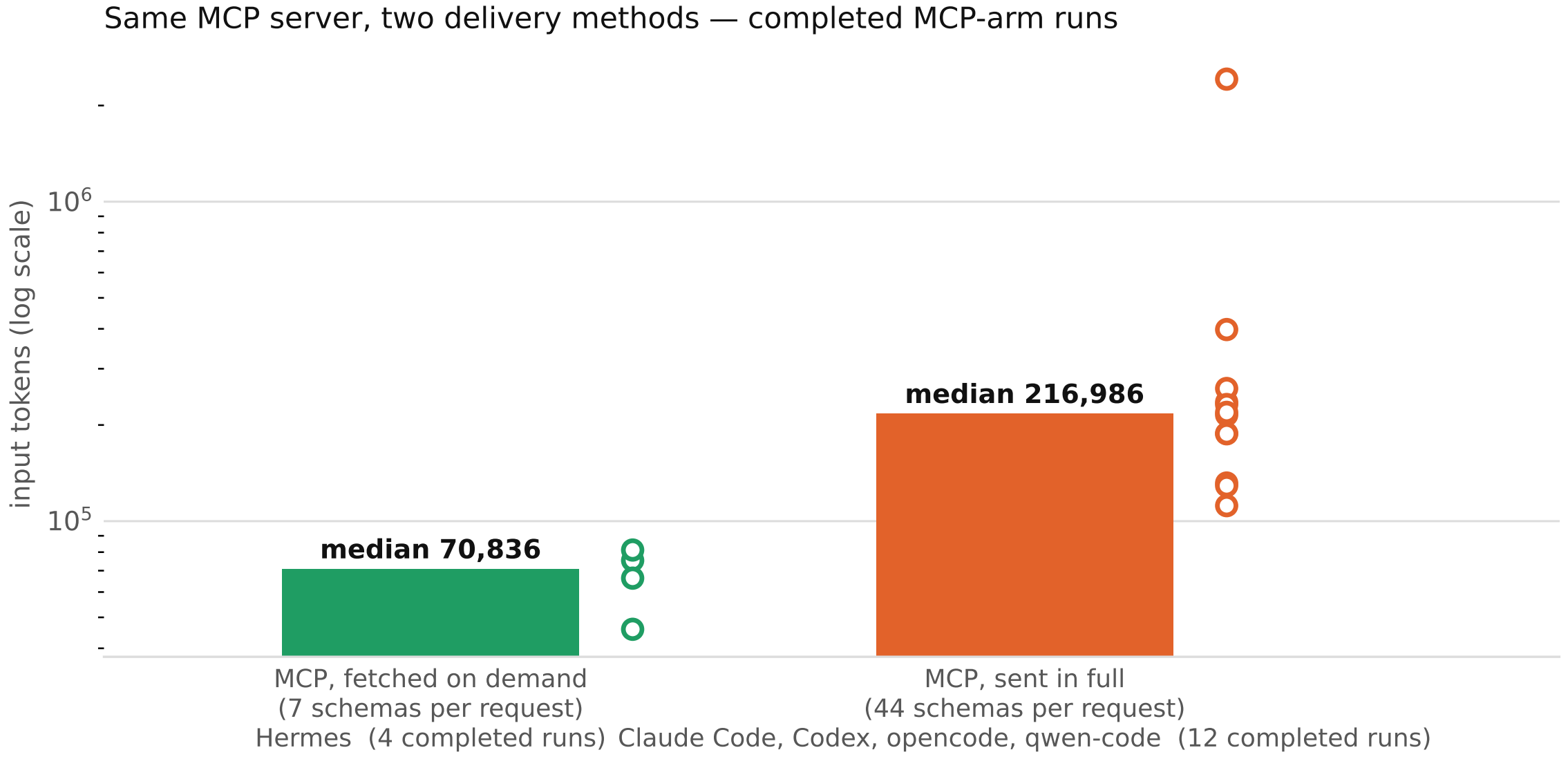

Figure 4: The same MCP server delivered two ways. Bars give the median input tokens of the MCP arm under each delivery method; circles are individual runs, on a logarithmic scale.

Codex, qwen-code, Claude Code and opencode all send the GitHub server's complete catalogue — all forty-four tool descriptions — with every request. A task that takes twenty exchanges transmits that catalogue twenty times. Claude Code sends seventy-four schemas in total on that arm: its own twenty-seven built-in tools, three for reading MCP resources, and the server's forty-four. Its command-line arm still carries the twenty-seven, which is why that arm begins around twenty-two thousand tokens before the task has said anything.

Hermes sends **seven** schemas. Two of those seven are a gateway pair: one that lists and describes the available tools, and one that invokes a tool by name. The remaining forty-two descriptions are fetched only when the model asks for them. The agent has access to the same forty-four operations; it does not carry their descriptions in every request.

**Table 5.** Cost of the MCP arm by delivery method, completed runs only. "Schemas per request" counts the tool descriptions from the MCP server that are transmitted with each request to the model. The runs cover whichever scaffolding–model pairs completed: four runs of Hermes against twelve runs spread over four scaffoldings.

| Delivery method | Schemas per request | Median tokens | Completed runs |
|---|---|---|---|
| Fetched on demand | 7 | 70,836 | 4 |
| Sent in full every request | 44 | 216,986 | 12 |

The difference is a factor of **3.1**, and it is a property of the agent scaffolding rather than of the protocol. Both delivery methods use MCP, both reach the same server, and both expose the same operations to the model. What differs is whether the scaffolding transmits the MCP catalogue eagerly or on request.

The direction is corroborated by the protocol's own authors. Anthropic [2026c] describe reducing MCP token consumption substantially by having the model write code that calls tools, rather than presenting every tool definition in the request — the same principle of deferring the tool descriptions until they are needed. That account is an engineering report rather than a controlled comparison, and it concerns a different mechanism from the gateway used here, but it establishes that eager transmission is a design choice rather than a requirement of the protocol.

This comparison should be treated with caution. Only one scaffolding in our sample fetches tool descriptions on demand, so the comparison in Table 5 sets four completed runs of one scaffolding against twelve completed runs of four others, and the delivery method is confounded with every other difference between Hermes and the rest. The measurement is solid — Hermes did use the same server, expose the same operations, and transmit a sixth as many schemas — but attributing the cost difference to delivery specifically requires a second scaffolding that delivers its MCP catalogue the same way, which we did not have. We report it as a mechanism worth testing rather than one we have established.

### 4.4 The cost of failure

The comparisons above exclude runs that failed, because a cost per completed task cannot be computed where no task was completed. Those runs still cost money — more of it, on average, than the runs that succeeded — and what they bought was nothing. Table 6 reports that directly, over every live run and with nothing excluded.

**Table 6.** Tokens and money spent on runs that did not complete the task. The first two rows cover the five MCP-capable scaffoldings; the third covers pi and Tau. "Wasted %" is the share of the group's tokens, and "Wasted cost %" the share of its modelled spend (Section 2.5), that produced no completed work.

| Group | Runs | Failed | Wasted tokens | Wasted % | Wasted cost % |
|---|---|---|---|---|---|
| Command-line arm | 19 | 3 | 234,675 | 7.3% | 2.2% |
| MCP arm | 19 | 3 | 542,232 | 9.8% | 12.9% |
| No MCP client | 8 | 0 | 0 | 0.0% | 0.0% |

Just under ten per cent of the tokens consumed by MCP arms, and **12.9 per cent of the money**, were spent on runs that delivered nothing. The corresponding figures for the command-line arm are 7.3 per cent and 2.2 per cent. The two scaffoldings with no MCP client wasted nothing at all, because they failed no runs.

**The two arms failed the same number of times.** Three runs of nineteen in each: the difference in the final column is entirely a difference in what those failures cost. The three command-line failures were cheap configurations — two runs of the 27-billion-parameter local model and one of the cheapest hosted model — totalling \$0.06. The three MCP failures were expensive ones, including two Codex runs and one on GLM, totalling \$0.41. A failed MCP run in this study cost about six times a failed command-line run.

This is not evidence that the protocol is less reliable; on these data it is not less reliable. It is evidence that when failure happened on the MCP side it happened on costly cells. Whether that is a property of the interface or an accident of which six runs failed, six runs cannot say.

What does not depend on any of this is the direction of the underlying cost: a failed run has no cost per unit of delivered work, the ratio being unbounded, and failure is the expensive outcome rather than the cheap one. The median failed run in this study consumed 170,937 input tokens against 83,600 for the median completed one — twice the spend for nothing.

The repetitions agree: twenty-five of twenty-nine completed in each arm, in data gathered weeks apart. The wasted-cost share is not a reliability figure, and equal failure counts do not make MCP free.

**Where the failures sit changes the reading, so we also split the analysis by model scale.** The split is post hoc — we chose it after seeing where the failures fell — but the splitting variable is not arbitrary: model scale is the axis Section 1.2 already argues should matter, and the smallest model is the one place the two arms behave differently.

Set the 27-billion-parameter model aside and the parity dissolves. On the remaining fourteen runs per arm, the command line failed once in fourteen — a reporting miss worth a cent — while the MCP arm failed three times in fourteen, and the wasted-cost shares move to 18.7 per cent against 0.3. The repetitions split the same way: without the smallest model the command-line arm completed every one of its fourteen measured runs, and the MCP arm twelve of fifteen.

The two groups of failures also differ in kind. The small model's failures are pairing artifacts, not evidence that the shell is beyond it: under the other five scaffoldings the same model completed nineteen of its twenty shell-only attempts, and under pi and Tau it did so at the lowest costs in the study. Where it failed it abandoned the workflow partway or skipped the closing report, and it completed every MCP-arm attempt those same two scaffoldings gave it (Section 6). The failures on capable models point the other way, and one of them repeats: qwen-code driving GLM-5.2 with the MCP server attached failed all four attempts across both datasets, the only configuration in the study that never completed. Codex's two hosted MCP runs also stalled at a quarter of the task, though each ran once and single runs prove little.

One reading of this split, offered as interpretation rather than result: failures that come from model capacity are transient, because small models improve on a cadence of months, while a failure that repeats under one scaffolding and one interface is structural and will not be fixed by a better model. On that reading the failures that persist will be the MCP side's, because better small models will remove the others. Our data cannot establish this; a study designed around model scale could.

Where failures are rare and unevenly priced, one or two runs dominate the wasted-cost share. This is why we repeat runs — and why, when a run failed because of an error in our setup or configuration, we fixed the error and repeated the run instead of scoring it.

**4.5 Dependence on turn count**

Because the MCP catalogue is retransmitted with every request, its cost is multiplied by the number of exchanges rather than paid once. Anything that causes an agent to take more turns therefore also multiplies what that catalogue costs. This is consistent with the extreme case in Section 3, where a model that repeated itself across 108 tool calls incurred a cost an order of magnitude beyond a typical MCP run rather than a proportionally small increase.

## 5. Results: cost by model

The five models differ in list price by a factor of twenty, so a comparison expressed only in tokens hides which choices are actually expensive. Table 7 reports each model over its completed runs, in tokens and in money.

| Model | Served | Completed | Tokens | Cost | Cost, cached discounted |
|---|---|---|---|---|---|
| `gpt-5.6-luna` | OpenAI | 8/10 | 134,618 | \$0.0141 | \$0.0035 |
| `qwen3.6:27b` | local | 10/12 | 97,890 | \$0.0339 | \$0.0339 |
| `gpt-5.6-terra` | OpenAI | 9/10 | 91,409 | \$0.1053 | \$0.0311 |
| `glm-5.2` | local | 11/12 | 102,031 | \$0.1243 | \$0.0178 |
| `sonnet-5` | Anthropic | 2/2 | 389,516 | \$0.8022 | \$0.1405 |

**Table 7.** Results by model, over runs that completed the task. "Completed" gives completed runs out of runs attempted. "Cost" prices the run at list rates; the final column applies a ninety per cent discount to cached input. The two locally served models are priced counterfactually, as what renting equivalent capacity would cost. `qwen3.6:27b` shows no difference between the two columns because the local server reports no cache figure at all.

Three observations follow.

**Token consumption is flat across models, and cost is not.** The median completed run consumes between 74,755 and 83,600 tokens for four of the five models, yet costs between \$0.0085 and \$0.0922 — a factor of eleven arising almost entirely from price per token rather than from tokens consumed.

`sonnet-5` is the exception on both dimensions, and it is also the only model tested exclusively under Claude Code, so its figures reflect that pairing rather than the model alone.

**The cheapest model completed the task.** `gpt-5.6-luna`, at $0.0085 per completed run, is an order of magnitude cheaper than `glm-5.2` or `gpt-5.6-terra` for the same work. Its completion rate, six of eight, is the weakest among the hosted models, which is the trade a practitioner would have to weigh.

**The open-weight models were the most reliable in the study.** `qwen3.6:27b` finished ten of twelve runs, at a modelled $0.0339 and on hardware a single workstation can host. Section 6 examines that case in more detail.

## 6. Results: open-weight models on local hardware

The two open-weight models we served on our own hardware — a 27-billion-parameter model and a larger mixture-of-experts model — are the kind a team would choose to run locally rather than pay an API provider for. Table 8 reports every run of the 27-billion-parameter model, sorted by cost.

**Table 8.** All runs of the 27-billion-parameter open-weight model. "Done" is completion over the four verification conditions. "Relative" expresses cost as a multiple of the cheapest configuration.

| Scaffolding | Arm | Tokens | Relative | Done |
|---|---|---|---|---|
| Tau | command line | 17,416 | ×1.0 | 4/4 |
| pi | command line | 25,548 | ×1.5 | 4/4 |
| Hermes | MCP | 66,320 | ×3.8 | 4/4 |
| opencode | command line | 66,448 | ×3.8 | 3/4 |
| Codex | command line | 83,247 | ×4.8 | 4/4 |
| Hermes | command line | 83,954 | ×4.8 | 4/4 |
| Claude Code | command line | 94,572 | ×5.4 | 1/4 |
| opencode | MCP | 111,826 | ×6.4 | 4/4 |
| Claude Code | MCP | 188,183 | ×10.8 | 4/4 |
| qwen-code | command line | 306,490 | ×17.6 | 4/4 |
| qwen-code | MCP | 397,922 | ×22.8 | 4/4 |
| Codex | MCP | 2,418,828 | **×138.9** | 4/4 |

**The model completed the task under every scaffolding**, and in ten of the twelve configurations it was given. It was rarely the limiting factor. What changed across the table is not whether the work got done but what it cost to do it, and that varied by a factor of 139.

Two configurations fall short, and both point the opposite way to Section 1.2's argument. Both are command-line arms, and in both cases the same scaffolding completed the task with the MCP server attached.

Under Claude Code's command-line arm the model abandoned the workflow after locating the issue, satisfying one condition of four, while its MCP arm completed three times out of three.

Under opencode's command-line arm it did almost everything — branch, patch and pull request all verified against the repository — and satisfied three conditions of four, missing only the closing report of the file count. Its MCP arm, again, completed every attempt.

Section 8.1 shows both command-line cells completing sometimes, so these are marginal pairings rather than flat incapacities. And the shell itself is not the difficulty: under the other five scaffoldings the same model completed nineteen of its twenty shell-only attempts, and under pi and Tau it did so at the lowest costs in the study. The model needed no saving from the command line — it needed saving from two particular scaffoldings' command-line arms, and within those two MCP is what happened to work. Section 1.2 predicts a small model should struggle with the MCP convention before it struggles with the shell; what these pairings show instead is that the scaffolding sits between the model and either interface, and can make even the familiar one fail. That is the conclusion Section 3 reaches by price, arrived at here by reliability. We report it as observations on one model and one task, not as a refutation.

The larger open-weight model also did its best work on the thinnest scaffoldings, and it adds the only failure in this group. It completed the task on every scaffolding except one — qwen-code with the MCP server attached, where it managed half the conditions — and its cost ranged from 13,588 tokens on pi to 322,868 on qwen-code, a factor of 24.

A team wanting to run agents on modest hardware, or on models they host themselves, is not primarily constrained by model capability for work of this size. They are constrained by how much the scaffolding spends on the model's behalf. A 27-billion-parameter model needing 17,416 tokens per task is a viable proposition on a single workstation; the same model needing 2.4 million is not, and the difference between those two situations is a software choice, not a hardware one.

## 7. Results: adherence to the assigned interface

Before the isolated design of Section 2.4 was adopted, we ran a companion experiment: the same task family against a public repository, with the GitHub MCP server attached in twenty-one runs and confirmed connected in each. Because the fixture was public and credentials were not separated, agents in that experiment had several routes to the repository — which made it useless for cost comparison and, for the same reason, informative about behaviour. We examined which tools each run actually called.

Six runs used the MCP tools exclusively. Six completed the task entirely through shell commands without calling an MCP tool once. Six used both. Three called no tools at all. Four bypassed both interfaces and queried the GitHub web API directly over HTTP.

Adding an instruction to the prompt that named the available interface and directed the agent to use it did not change this behaviour materially.

Configuration proved more effective than instruction. Once we isolated the arms as described in Section 2.4, removing the credentials that made the alternative route possible, all twenty-five command-line runs in the main matrix whose tool calls were recorded used only the command line. The MCP arm is less tidy: of its seventeen runs with a recorded register, eleven used the MCP tools and nothing else, three of Codex's runs mixed shell commands into it, one of opencode's reached for a built-in note-keeping tool alongside it, and two never called an MCP tool at all.

This has a consequence for how such comparisons should be read. **A measurement that assigns an interface without verifying which interface was used reports the cost of an unknown mixture.**

The bias runs in a predictable direction: an agent that quietly finishes the task through its shell while an unused MCP catalogue sits in its context produces a low “MCP” measurement, so studies that do not check understate what MCP costs.

## 8. Limitations

### 8.1 A replication, and how much a single run can be trusted

Every figure reported above rests on one run per configuration. To establish what that is worth, we repeated every configuration using a locally served model three times, changing nothing between repetitions. Local inference has no marginal cost, so the price of this was time. The repetitions produced 70 executed runs, 61 of which completed the task, and they are published alongside the main dataset.

**How much does an identical configuration vary?** For each configuration with more than one completed repetition, we divided the largest token count by the smallest.

| | Spread, largest ÷ smallest |
|---|---|
| Median across configurations | **1.51×** |
| Widest single configuration | 5.41× |

Across twenty-one configurations the typical spread is a factor of 1.5. This is the resolution limit of the study: **a difference smaller than roughly twofold cannot be distinguished from run-to-run variation**. Nine of the thirteen paired ratios of Section 4.1 fall between 0.61 and 1.30, inside that band. Four lie beyond it, and not all in the same direction: two Codex pairs at 2.27 and 29.06 and one opencode pair at 1.79, where MCP cost more, and one Claude Code pair at 0.43, where it cost less than half. Outliers on both sides are the bimodality that makes the pooled verdict inconclusive rather than null. The differences the paper does rely on — the 20-fold span between scaffoldings in Section 3 and the 139-fold gap in Section 6 — lie far outside the band.

**Does the principal result reproduce?** The repetitions were run after the analysis above was complete, so they function as a test of it rather than as input to it.

| Condition | Completed |
|---|---|
| Command-line arm | 25/29 |
| MCP arm | 25/29 |
| Scaffoldings with no MCP client | 11/12 |

**The two arms complete at the same rate here: twenty-five of twenty-nine each.** That agrees with the main matrix, where the arms also failed equally often, and it sharpens the point of Section 4.4: the failure story is about money, not frequency. The split by model scale reproduces here as well. Without the smallest model the command-line arm completed every one of its fourteen measured runs while the MCP arm completed twelve of fifteen — and all three of those failures are the same configuration, qwen-code driving GLM-5.2, which never completed in either dataset.

What survives the replication is the *cost* asymmetry rather than the *rate* one. Even with the arms tied on completion, the money lost to failure remains uneven, because the runs that failed were the expensive ones.

**One qualification.** The single failure among the two scaffoldings without MCP was Tau, on the 27-billion-parameter model. It opened a pull request and reported the file count correctly while never creating a branch or applying the patch: two of four conditions satisfied, eight tool calls, no error raised, and a clean exit. Tau is an educational implementation and this is the smallest model in the study, so an occasional lapse in that pairing is unremarkable; across both datasets it amounts to one failure in twenty attempts. It tempers the claim that these scaffoldings never failed, and it is precisely the failure mode that verification by repository inspection exists to catch — an agent that reports success and leaves no trace would pass any check that read only its own account of itself.

**Hosted configurations are largely single-run.** The repetitions of Section 8.1 cover the locally served models, which cost nothing to repeat. Hosted configurations were run once, for reasons of budget rather than principle, with one exception: the Claude Code MCP cell was repeated three times when it was re-run after the credential fault of Section 2.6, and varied by a factor of 1.19. One hosted configuration — Hermes on an OpenAI model, command-line arm — happened to be run three times in the course of debugging and varied by a factor of 3.6, so the resolution limit of Section 8.1 applies to the hosted rows at least as strongly as to the local ones.

**One task, in one domain.** The task exercises GitHub, a service with both a mature command-line client and an official MCP server. A domain where the MCP offers an operation with no convenient CLI equivalent would plausibly favour MCP more than this task does.

**How the scaffoldings were chosen.** They were selected for prominence and for meeting four requirements the design imposes: the scaffolding must run as a terminal program that a script can invoke as a subprocess, offer a non-interactive mode, accept an arbitrary model endpoint so the model can be held fixed, and ship an MCP client — the last except for the two controls, which are included precisely because they do not. Those requirements exclude two prominent tools on principle rather than by oversight: Cline is a code-editor extension and OpenHands a containerised platform, and neither can be driven the way this harness drives the others.

Prominence we measure by GitHub stars, which is a popularity proxy and nothing better, but is public and dated. On that measure the seven span roughly the top of the field: Hermes, opencode, Claude Code and Codex are four of the five most-starred, pi is more popular than several tools we did not run, and Tau is small and included for a stated reason — it is an independent implementation of pi's design, which is what allows Section 3 to claim the minimal result survives reimplementation rather than being a property of one codebase.

Three scaffoldings appear to meet the same four requirements and were not run: `gemini-cli`, `goose` and `crush`. Time rather than principle is the reason. The sample should be read as a purposive one covering widely used tools, not as representative of the population, and the selection was made before we counted stars rather than derived from the count.

**Uneven coverage.** Claude Code contributes six live runs to the main matrix against eight for most others, and two of the six are on a subscription credential that does not cross the proxy. Its row in Table 1 therefore rests on a smaller sample, and readers who wish to exclude the two self-reported cells can do so: the twelve proxied runs behind Section 8.1 agree with the scaffolding's own accounting to the token, so nothing in the row depends on trusting it. Hermes runs on OpenAI models did not report which tools were called, for reasons internal to how those responses are transmitted, so Hermes contributes cost and completion figures but is excluded from the analysis of tool-call behaviour in Section 7.

**The delivery-method comparison rests on one scaffolding.** Section 4.3 compares four completed runs of the single scaffolding that fetches tool descriptions on demand against twelve completed runs of four scaffoldings that transmit them in full. Delivery method is therefore confounded with every other difference between that scaffolding and the others, and no amount of care in the measurement separates them. A second scaffolding using on-demand delivery would be required to test the mechanism, and none was available to us.

**The control scaffoldings differ in more than one respect.** pi and Tau lack MCP support, but they differ from the other four in ways we did not measure. Their advantage in Table 1 cannot be attributed to the absence of MCP alone. Section 4.3 offers a mechanism that makes the attribution plausible without establishing it: the one MCP-capable scaffolding that does not transmit its MCP catalogue eagerly is also the cheapest MCP-capable scaffolding by a factor of two.

## 9. Discussion

### 9.1 Summary of findings and their evidential basis

Our results support four statements of differing strength; a reader may accept some and reject others.

**An MCP catalogue is not necessary for this class of work.** Two scaffoldings with no MCP client completed every run using ordinary command-line tools. This is the most robust claim in the paper: it requires only that the runs happened, and the verification in Section 2.2 confirms they did. Its scope is the class of task we tested — operations against a service that already has a mature command-line client — and it does not extend to services with no such client.

**Scaffoldings without an MCP client substantially outperformed five widely used ones.** On the like-for-like comparison in Section 3.2, with no MCP catalogue attached to anything, the difference ranges from 5.0× to 28× in cost, and the two scaffoldings without MCP completed every run in the main matrix. This is a comparison between two populations of scaffolding rather than an isolation of one feature.

**Within a scaffolding that supports both, the arms separate on the cost of failure and not on its frequency.** Section 4.1 finds thirteen strictly paired ratios spanning 0.43 to 29.06 — unresolvable against the measured run-to-run variation, so the conventional cost comparison is inconclusive. What does separate the arms is Section 4.4: 12.9 per cent of the money spent on the MCP arm bought no completed work, against 2.2 per cent on the command line. But the arms failed the same number of times — three of nineteen each in the main matrix, twenty-five of twenty-nine each in the repetitions — so what we are reporting is that the MCP arm's failures were the expensive ones, not that it failed more often. The rate parity itself is carried by the smallest model: set it aside, per the split in Section 4.4, and both the rate and the cost of failure sit on the MCP side — including the one configuration in the study that never completed at all, an MCP arm. Both are within-scaffolding observations, so they isolate the interface; the cost figure is the one that speaks to the question as it is usually posed. It is also the figure most exposed to a single bad cell, as Section 4.4 records, so it should be read to one significant figure and no further.

**Agents cannot be relied upon to use the interface they are assigned.** Section 7 reports that six of twenty-one runs used the attached MCP tools exclusively. Prompting did not fix it; removing the alternative did.

**9.2 Implications for agent construction**

The two scaffoldings without MCP clients are both open source, small enough to read in an afternoon, and licensed for modification. Taken with the first two claims above, this matters for anyone building a purpose-specific agent rather than adopting a general-purpose assistant.

The five widely used scaffoldings in our sample are built for open-ended work: they carry the machinery that breadth requires, and they carry it on every task including tasks that need almost none of it. Our task needed a shell, a few file operations and a credential. On that task the two scaffoldings without MCP were between five and twenty-eight times cheaper and failed less often, and both can be modified to fit a specific job.

For a well-specified, repeatable task — the kind that runs in a pipeline, or on a schedule, or many times a day — the evidence favours a small scaffolding that a team controls over a general assistant that a team configures. A general-purpose assistant remains appropriate for open-ended exploratory work, where breadth is what the user is paying for. The same split applies to MCP itself: for an agent that cannot know in advance which services it will meet, a discoverable, standardised catalogue earns its cost; for an agent whose operations are known when it is built, the catalogue describes work the agent will never do. This is the outcome the harvest step of *Agents All the Way Down* predicts [Alier et al. 2026a]: once the task boundary is known, the scaffolding can be reduced to fit it, and on this task the reduction was worth orders of magnitude more than the choice of interface.

**9.3 Implications for self-hosted models**

Section 6 showed that a 27-billion-parameter model completed our task under every scaffolding tested, at costs varying by a factor of 139. The model was never the constraint; the software around it was.

The gap widens as the model gets smaller, which sharpens the argument of Section 9.2: the five general-purpose scaffoldings carry the machinery of breadth on every task, and a small model has the least context to spare for it. This is not evidence that general-purpose assistants are poorly built; it is evidence that generality has a running cost, paid per task whether or not the generality is used, and that small models pay it at the highest rate.

Taken together with the per-model costs of Section 5, this suggests a route to running agents on modest hardware that does not depend on waiting for better small models. The models we tested were already sufficient for this task. What made them expensive, or affordable, was the scaffolding around them.

**9.4 Delivery method as a candidate mechanism**

A single scaffolding is not enough to attribute the cost difference of Section 4.3 to delivery method, and we do not do so. The claim we can support is narrower: the number of schemas transmitted per request varies by a factor of six between scaffoldings using the same protocol, and this quantity is not reported in any published comparison we are aware of. Whether it explains the cost difference is a question for a study designed around it.

**9.5 Recommendations for future measurement**

Three practices would make published comparisons easier to reconcile. The first is to verify which interface an agent actually used rather than which it was given, since Section 7 shows the two differ often enough to invalidate the assumption. The second is to report the number of tool schemas transmitted per request. That quantity varies by a factor of six between scaffoldings using the same protocol, which means two studies can measure the same protocol and the same server and still not be comparing the same thing. Whether it accounts for the disagreement in existing estimates is untested, but it cannot be ruled in or out while nobody reports it.

The third is to separate failures of the thing under test from failures of the apparatus, and to re-run the second kind rather than score them. A benchmark of this shape will produce runs that die on a rejected credential, an unset path or an unreachable host, and those runs say nothing about the interface being compared. Scoring them as failures publishes the experimenter's own misconfiguration as a property of the protocol.

Our Section 2.6 cell is the worked example. Read as a failure it put the wasted-cost share above forty per cent; re-run with a working credential it completes, and the figure is 12.9. The fault was reported in the agent's closing message, but nothing in the run's cost, its completion score or its tool calls would have prompted anyone to look. A study that reports its failures without saying which kind they were is asking to be read as more conclusive than it is.

The task definition, the verification code, the measurement harness and every individual run are published alongside this paper, so that these results can be checked — and contradicted — using the same instruments that produced them.

## Appendix A — every individual run

The tables below list all 54 runs, and are generated from the published dataset by `bench/appendix_v9.py` rather than maintained by hand. **Tokens** is input-token consumption. **Cache** is the share of that input the provider billed at the reduced cached rate. **Calls** is the number of tool invocations recorded; a dash means the agent scaffolding did not report them in a form our proxy could read. **Schemas** is how many tool descriptions were transmitted per request, where the scaffolding disclosed it. **Done** is completion over the four verification conditions of Section 2.2.

Cells marked *void* were never executed: pi and Tau have no MCP client, so their MCP arm cannot exist. They are listed rather than omitted, because "cannot be run" and "costs nothing" are different statements and the table should keep them apart. No cell is omitted for having failed — including the one whose MCP server could not authenticate, which appears here with the figures from its re-run.

Model names are abbreviated: luna and terra are `gpt-5.6-luna` and `gpt-5.6-terra`, qwen27b is `qwen3.6:27b`, sonnet is `sonnet-5`.

**MCP-capable harnesses**

| Harness | Model | Arm | Tokens | Cache | Calls | Schemas | Done |
|---|---|---|---|---|---|---|---|
| claude | glm-5.2 | cli | 278,455 | 100% | 13 | 27 | 100% |
| claude | glm-5.2 | mcp | 260,170 | 99% | 9 | 74 | 100% |
| claude | qwen27b | cli | 94,572 | 0% | 3 | 27 | 25% |
| claude | qwen27b | mcp | 188,183 | 0% | 6 | 74 | 100% |
| claude | sonnet | cli | 543,139 | 97% | 14 | — | 100% |
| claude | sonnet | mcp | 235,892 | 88% | 8 | — | 100% |
| codex | glm-5.2 | cli | 102,031 | 99% | 20 | — | 100% |
| codex | glm-5.2 | mcp | 231,797 | 93% | 30 | — | 100% |
| codex | luna | cli | 81,510 | 88% | 14 | — | 100% |
| codex | luna | mcp | 189,489 | 87% | 10 | — | 25% |
| codex | terra | cli | 39,378 | 73% | 6 | — | 100% |
| codex | terra | mcp | 181,806 | 86% | 10 | — | 25% |
| codex | qwen27b | cli | 83,247 | 0% | 16 | — | 100% |
| codex | qwen27b | mcp | 2,418,828 | 0% | 108 | — | 100% |
| hermes | glm-5.2 | cli | 74,755 | 96% | 15 | 6 | 100% |
| hermes | glm-5.2 | mcp | 45,875 | 82% | 12 | 7 | 100% |
| hermes | luna | cli | 73,655 | 88% | — | 6 | 75% |
| hermes | luna | mcp | 75,352 | 86% | — | 7 | 100% |
| hermes | terra | cli | 91,409 | 90% | — | 6 | 100% |
| hermes | terra | mcp | 81,222 | 89% | — | 7 | 100% |
| hermes | qwen27b | cli | 83,954 | — | 14 | 6 | 100% |
| hermes | qwen27b | mcp | 66,320 | — | 14 | 7 | 100% |
| opencode | glm-5.2 | cli | 73,486 | 99% | 11 | 10 | 100% |
| opencode | glm-5.2 | mcp | 131,649 | 98% | 9 | 57 | 100% |
| opencode | luna | cli | 241,612 | 93% | 21 | 10 | 100% |
| opencode | luna | mcp | 214,898 | 91% | 12 | 57 | 100% |
| opencode | terra | cli | 137,800 | 93% | 17 | 10 | 100% |
| opencode | terra | mcp | 129,114 | 86% | 9 | 57 | 100% |
| opencode | qwen27b | cli | 66,448 | — | 9 | 10 | 75% |
| opencode | qwen27b | mcp | 111,826 | — | 6 | 57 | 100% |
| qwen | glm-5.2 | cli | 322,868 | 99% | 11 | — | 100% |
| qwen | glm-5.2 | mcp | 170,937 | 97% | 8 | — | 50% |
| qwen | [illegible] | [illegible] | [illegible] | [illegible] | [illegible] | — | 100% |

**Harnesses with no MCP client**

| Harness | Model | Arm | Tokens | Cache | Calls | Schemas | Done |
|---|---|---|---|---|---|---|---|
| pi | glm-5.2 | cli | 13,588 | 96% | 6 | — | 100% |
| pi | glm-5.2 | mcp | void | | | | |
| pi | luna | cli | 15,731 | 87% | 13 | — | 100% |
| pi | luna | mcp | void | | | | |
| pi | terra | cli | 10,892 | 88% | 7 | — | 100% |
| pi | terra | mcp | void | | | | |
| pi | qwen27b | cli | 25,548 | 0% | 10 | — | 100% |
| pi | qwen27b | mcp | void | | | | |
| tau | glm-5.2 | cli | 15,502 | 86% | 7 | — | 100% |
| tau | glm-5.2 | mcp | void | | | | |
| tau | luna | cli | 11,554 | 77% | 6 | — | 100% |
| tau | luna | mcp | void | | | | |
| tau | terra | cli | 27,412 | 76% | 11 | — | 100% |
| tau | terra | mcp | void | | | | |
| tau | qwen27b | cli | 17,416 | — | 7 | — | 100% |
| tau | qwen27b | mcp | void | | | | |